\documentclass[graybox]{svmult}

\usepackage{mathptmx}       
\usepackage{helvet}         
\usepackage{courier}        
\usepackage{type1cm}        
\usepackage{makeidx}         
\usepackage{graphicx}        
\usepackage{multicol}        
\usepackage[bottom]{footmisc}
\usepackage{algorithm} 		 
\usepackage[noend]{algpseudocode} 
\usepackage{amsmath}
\usepackage{amssymb}
\usepackage{color}
\usepackage{cancel}
\usepackage{soul}
\usepackage{cite}

\definecolor{mygray}{RGB}{200, 200, 200}
\definecolor{mygreen}{RGB}{1, 200, 1}

\usepackage[absolute,overlay]{textpos}

\makeindex             

\begin{document}

\begin{textblock*}{20cm}(2.9cm,1.5cm) 
   \color{blue}\Large{\textsf{\textbf{International Symposium on Experimental Robotics (ISER 2018)}}} \color{black}
\end{textblock*}

\title*{Interactive Learning with Corrective Feedback for Policies based on Deep Neural Networks}
\titlerunning{Interactive Learning with Corrective Feedback for Policies based on DNNs}
\author{Rodrigo P\'{e}rez-Dattari, Carlos Celemin, Javier Ruiz-del-Solar, and Jens Kober}
\institute{Rodrigo P\'{e}rez-Dattari \at Universidad de Chile, Av. Tupper 2007, Santiago, Chile, \email{rodrigo.perez.d@ing.uchile.cl}
\and Carlos Celemin \at Delft University of Technology, Mekelweg 2, Delft, Netherlands, \email{c.e.celeminpaez@tudelft.nl}
\and Javier Ruiz-del-Solar \at AMTC Center, Universidad de Chile, Av. Tupper 2007, Santiago, Chile, \email{jruizd@ing.uchile.cl}
\and Jens Kober \at Delft University of Technology, Mekelweg 2, Delft, Netherlands, \email{j.kober@tudelft.nl}}

\maketitle

\abstract{Deep Reinforcement Learning (DRL) has become a powerful strategy to solve complex decision making problems based on Deep Neural Networks (DNNs). However, it is highly data demanding, so unfeasible in physical systems for most applications. In this work, we approach an alternative Interactive Machine Learning (IML) strategy for training DNN policies based on human corrective feedback, with a method called Deep COACH (D-COACH). This approach not only takes advantage of the knowledge and insights of human teachers as well as the power of DNNs, but also has no need of a reward function (which sometimes implies the need of external perception for computing rewards). We combine Deep Learning with the COrrective Advice Communicated by Humans (COACH) framework, in which non-expert humans shape policies by correcting the agent's actions during execution. The D-COACH framework has the potential to solve complex problems without much data or time required. Experimental results validated the efficiency of the framework in three different problems (two simulated, one with a real robot), with state spaces of low and high dimensions, showing the capacity to successfully learn policies for continuous action spaces like in the Car Racing and Cart-Pole problems faster than with DRL.}


\section{Introduction}
\label{sec:1}
Deep Reinforcement Learning (DRL) has obtained unprecedented results in decision-making problems, such as playing Atari games \cite{Mnih2013}, or beating the world champion in GO \cite{Silver2016}. Nevertheless, in robotic problems, DRL is still limited in applications with real-world systems \cite{Gu2017}. Most of the tasks that have been successfully addressed with DRL have two common characteristics: 1) they have well-specified reward functions, and 2) they require large amounts of trials, which means long training periods (or powerful computers) to obtain a satisfying behavior. These two characteristics can be problematic in cases where 1) the goals of the tasks are poorly defined or hard to specify/model (reward function does not exist), 2) the execution of many trials is not feasible (real systems case) and/or not much computational power or time is available, and 3) sometimes additional external perception is necessary for computing the reward/cost function. 

On the other hand, Machine Learning methods that rely on transfer of human knowledge, Interactive Machine Learning (IML) methods, have shown to be time efficient for obtaining good performance policies and may not require a well-specified reward function; moreover, some methods do not need expert human teachers for training high performance agents \cite{akrour2011preference,Knox:2009:ISA:1597735.1597738,Celemin2018AnInteractive}. In previous years, IML techniques were limited to work with low-dimensional state spaces problems and to the use of function approximation such as linear models of basis functions (choosing a right basis function set was crucial for successful learning), in the same way as RL. But, as DRL have showed, by approximating policies with Deep Neural Networks (DNNs) it is possible to solve problems with high-dimensional state spaces, without the need of feature engineering for preprocessing the states. If the same approach is used in IML, the DRL shortcomings mentioned before can be addressed with the support of human users who participate in the learning process of the agent.

This work proposes to extend the use of human corrective feedback during task execution to learn policies with state spaces of low and high dimensionality in continuous action problems (which is the case for most of the problems in robotics) using deep neural networks.

We combine Deep Learning (DL) with the corrective advice based learning framework called COrrective Advice Communicated by Humans (COACH) \cite{Celemin2018AnInteractive}, thus creating the Deep COACH (D-COACH) framework. In this approach, no reward functions are needed and the amount of learning episodes is significantly reduced in comparison to alternative approaches. D-COACH is validated in three different tasks, two in simulations and one in the real-world.

\section{Related Work}
\label{sec:3}
This paper proposes a novel alternative to adapt policies, combining IML and DL. Specifically, we focus on techniques which transfer the teacher's knowledge based on occasional human feedback that may be either evaluative or corrective. Evaluative feedback has been used similarly to RL in methods wherein a human teacher communicates the desirability of the executed action or policy, with validations in problems of state spaces of either low dimensionality  \cite{Knox:2009:ISA:1597735.1597738,akrour2011preference} or high dimensionality \cite{Christiano2017,Warnell2017}. In contrast, corrective feedback is given by the teacher directly in the action's domain in order to modify the magnitude computed by the policy. To the best of our knowledge, corrective feedback has been only validated in problems with state spaces of low dimensionality \cite{Celemin2018AnInteractive,Argall2008}.
%
\section{Deep COACH}
\label{sec:4}
With COACH, a human teacher can advise a correction signal to the actions that the agent executes. If the agent executes an action $a$ that the human considers to be erroneous, then s/he would indicate the direction in which the action should be corrected (increment or decrement); thus, COACH was proposed for problems with continuous actions. Each dimension of the action would have a corresponding correction signal $h$ with values $0$, $-1$ or $1$ which produces an error signal with arbitrary magnitude $e$ that is used to directly shape the policy in a supervised manner. Thus, $\mathit{error}=h \cdot e$, where $h=0$ indicates that no correction has been advised. $h=\pm 1$ indicates the direction of the advised correction.

In this framework, we use two types of neural network architectures: feed forward fully-connected (FNN) for low-dimensional state problems, and convolutional neural networks (CNN) for high-dimensional state problems, e.g., raw image state spaces. In both cases the policies are updated every time feedback is received and also by sampling from a memory buffer $B$ with a fixed frequency every $b$ time steps. Every time the user advises a correction, the buffer $B$ is fed with the current state and a label generated by adding the action taken with the error correction $y_{label}=a+\mathit{error}$. In the case of the CNN architecture, the convolutional layers are trained offline before the interactive process for learning a low-dimensional representation of the state. The state is embedded in the latent space of an autoencoder  trained with a database of the agent exploring the  environment. In Algorithm \ref{algorithm:DeepCOACH}, the pseudocode of D-COACH is presented.

\begin{algorithm}[t]
\caption{D-COACH }\label{algorithm:DeepCOACH}
\begin{algorithmic}[1]
\State \textbf{Require:} error magnitude $e$, buffer update interval $b$, buffer sampling size $N$, buffer size $K$, pre-trained encoder parameters (if convolutional) 
\State \textbf{Init:} $B = []$  \emph{\# initialize memory buffer}
\For{t = 1,2,...}{}
\State \textbf{observe} state $s_{t}$
\State \textbf{execute} action $a_{t}=\pi(s)_{t}$
\State \textbf{feedback} human corrective advice $h_{t}$
\If{$h_{t}$ is not \textbf{0}}
\State $\mathit{error}_{t} = h_{t}\cdot e$
\State $y_{label(t)} = a_{t} + \mathit{error}_{t}$ 
\State \textbf{update} $\pi(s)$ using SGD with pair ($s_{t}$, $y_{\mathit{label}(t)}$) 
\State \textbf{update} $\pi(s)$ using SGD with a mini-batch sampled from $B$
\State \textbf{append} $(s_{t}, y_{\mathit{label}(t)})$ to $B$
\If{length($B$) $> K$ }
\State $B = B[2:K+1]$
\EndIf
\EndIf
\If{mod(t, b) is 0 and $B$ is not $\emptyset$}
\State \textbf{compute} $\pi(s)$ using SGD with a mini-batch sampled from $B$
\EndIf
\EndFor
\end{algorithmic}
\end{algorithm}

In the original COACH, it is proposed that each dimension should be trained independently \cite{Celemin2017}, which has the advantage of creating a working framework that does not need any prior information about the problem in order to give corrections. We call this type of policy updating \emph{decoupled} training, so a correction in an specific action dimension does not modify the magnitude of the actions in other axes for the same corresponding state. However, in this work we consider that for some problems it may be advantageous to exploit prior user knowledge about relations between the different dimensions of the actions. In this way, a correction in one of the action axes may be used to update more than one dimension. We call this case \emph{coupled} training.

\section{Experiments and Results}
Our proposed algorithm is validated experimentally in three different problems: (i) Cart-Pole (continuous action), which is a simulated task with low-dimensional state space; (ii) Car Racing, a simulated task with high-dimensional state space (raw pixels of an image); and (iii) Duckie Racing, a task with a real robot featuring a high-dimensional state space (raw pixels of an image). 

The experiments with the simulated agents are intended to compare the complete D-COACH presented in Algorithm \ref{algorithm:DeepCOACH}, along with a version of it without buffer (ignoring lines 2 and 11-16), and with a well known DRL agent (Deep Deterministic Policy Gradient DDPG \cite{Lillicrap2015} implemented by OpenAI \cite{baselines}). The comparison is carried out by plotting the cumulative reward obtained at each episode by the agent as a function of time. In the case of D-COACH, the obtained reward is only used as a performance metric. Also, the results are presented as a function of time instead of episodes (except in the Duckie Racing experiment), because episodes can have variable duration depending on the policy. Hence, the episode scale would not properly show the time taken by the learning process, which is an important characteristic, since D-COACH is meant to work with real robots. The simulated environments, Cart-Pole and Car Racing, were ran at $22.5$ and $20.5$ FPS, respectively. These experiments were carried out using human teachers and simulated teachers. Humans had approximately 5 minutes to practice teaching in each environment. The learning curves of agents trained by 10 human teachers were obtained and averaged; the learning curves of agents trained by a simulated teacher were repeated 30 times and averaged. Along with the algebraic mean, the confidence intervals that represent the $60^{th}$ percentile of the data were plotted. In the case of the Car Racing problem, it was observed that coupled training was advantageous when the teachers were humans. The designed coupled signals are shown in Table \ref{table:coupled_car_racing}.

\begin{table}[t]
\centering
\caption{Values of $h$ in the Car Racing problem for human teachers. When feedback is given, the generated correction acts over more than one dimension of the action. For instance, the feedback signal \emph{forward} means that the agent should simultaneously increase its acceleration and decrease its brake.}
\label{table:coupled_car_racing}
\begin{tabular}{lc}
\textbf{Feedback            } & \multicolumn{1}{l}{          }{\textbf{h
(direction, acceleration, brake)}} \\ \hline \hline
Forward     & (0, 1, -1)                                       \\ \hline
Back        & (0, -1, 1)                                       \\ \hline
Left        & (-1, -1, 0)                                      \\ \hline
Right       & (1, -1, 0)                                       \\ \hline
\end{tabular}
\end{table}

\begin{figure}[t]
    \centering
    \includegraphics[width=1.0\linewidth]{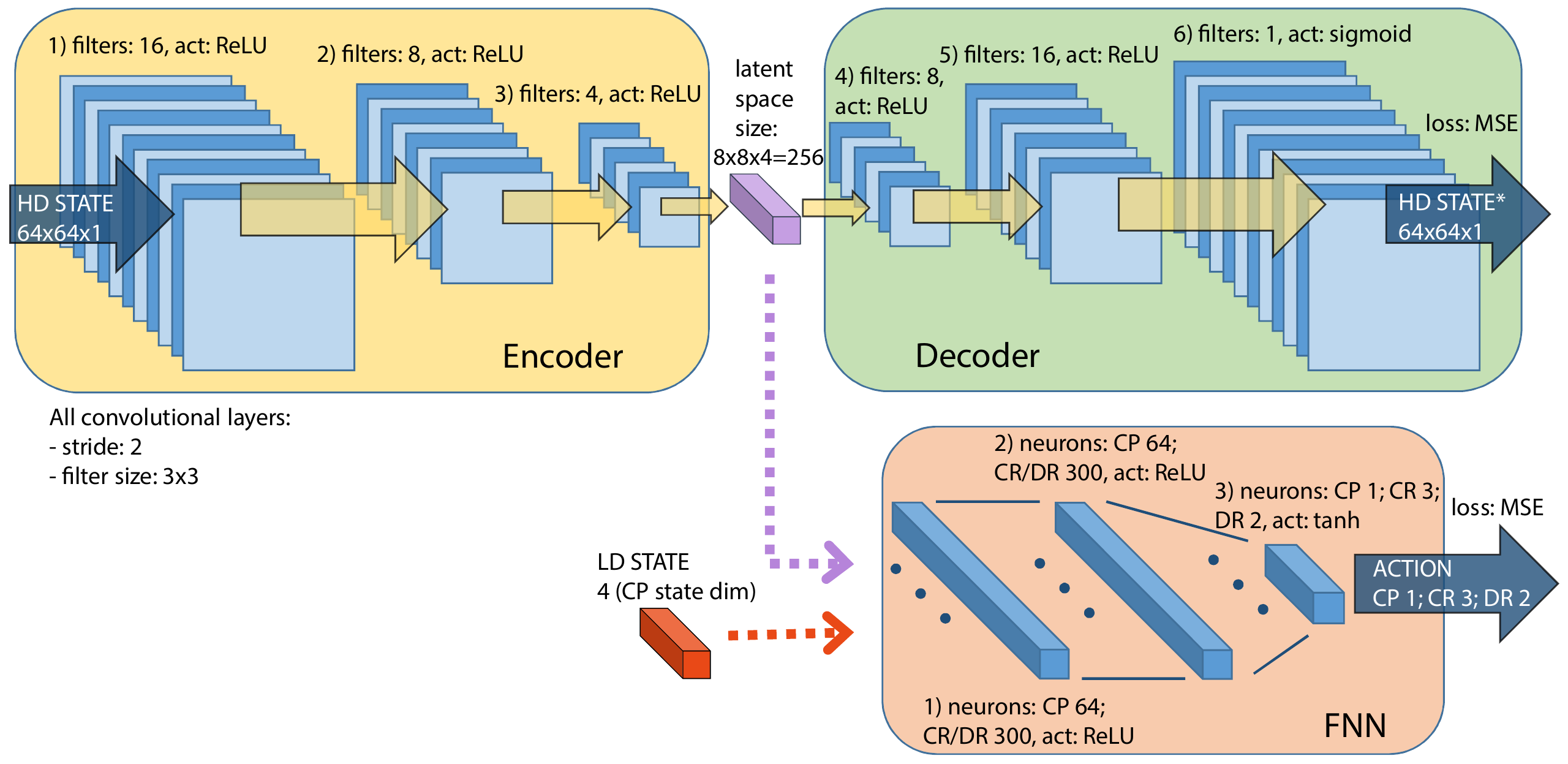}
    \caption{D-COACH neural networks architecture. Variations between environments are specified with the acronyms CP (Cart-Pole), CR (Car Racing) and DR (Duckie Racing). HD STATE: high-dimensional state space. LD STATE: low-dimensional state space.}
    \label{fig:network_diagram}
\end{figure}

The hyper-parameters of the neural networks used in these experiments were tuned with preliminary experiments. Different combinations of them were tested by a human teacher and the ones that made the training easiest were selected (see \figurename~{\ref{fig:network_diagram}}). The D-COACH error magnitude constant $e$ was set to $\textbf{1}$ in this paper.

\vspace{-0.3cm}

\subsection{Validation of replay buffer with simulated teachers}
The use of experience replay has been extensively validated in DRL; however, in this approach, we  still consider it necessary to test its impact. Unlike DRL, where the policy is updated with information collected from every time step, in COACH-like methods there only is new data to update the policy when feedback is given by the teacher, so the amount of data used to update the policy may be lower than in the RL case. Since the original COACH has been widely validated with real human teachers in several tasks, we carried out most of the comparisons using  a simulated teacher (a high performance policy standing-in as teacher, which was actually trained with D-COACH and a real human teacher) in this work, like in some of the experiments presented in \cite{Celemin2018AnInteractive}, in order to compare the methods under more controlled conditions. 

The simulated teacher generates feedback using $h = \operatorname{sign}(a_{\mathit{teacher}} - a_{\mathit{agent}})$, whereas the decision of advising feedback at each time step is given by the probability $P_{h} = \alpha \cdot\exp(-\tau\cdot \mathit{timestep})$, where $\{\alpha \in {\rm I\!R}\ | 0 \le \alpha \le 1\}$ and $\{\tau \in {\rm I\!R}\ | 0 \le \tau\}$. Additionally, since human teachers occasionally advise wrong corrections, a probability of giving erroneous feedback $P_{\mathit{err}}$ is added to the model. The variable $P_{\mathit{err}}$ indicates the probability that at least one dimension of $h$ is multiplied by $-1$ when feedback is given.

\begin{figure}[t]
    \centering
    \vspace{-0.2cm}
    \includegraphics[width=0.7\linewidth]{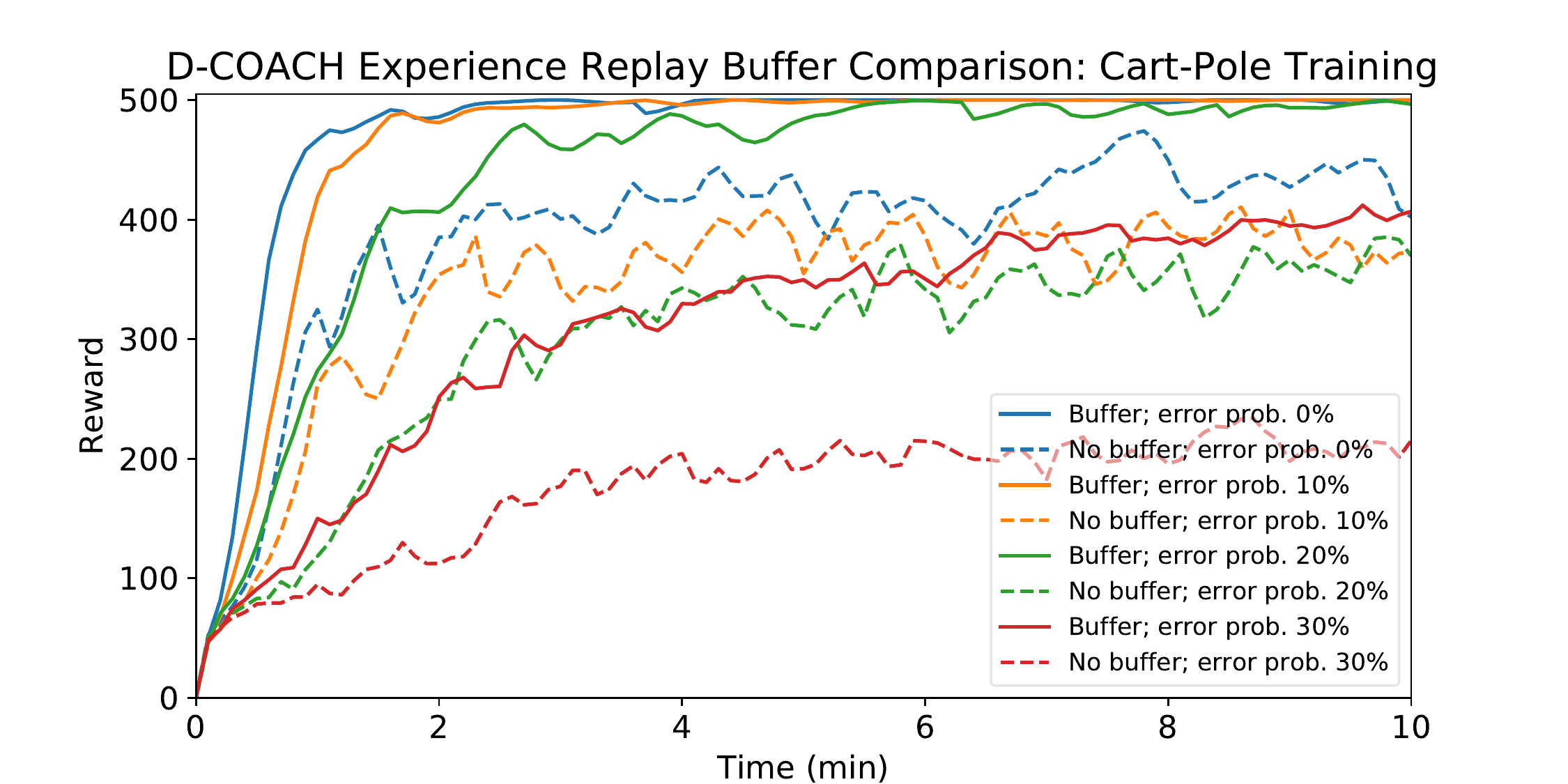}
    \vspace{-0.2cm}
    \caption{Comparison between using or not experience replay buffer for different values of $P_\mathit{err}$ in the Cart-Pole problem. Buffer: $K = 200$; $b = 10$; $N = 50$. $P_{h}$: $\alpha = 0.6$; $\tau = 0.0003$. Simulated teacher network learning rate: $0.0003$.}
    \label{fig:buffer_cart_pole}
\end{figure}

\begin{figure}[t]
    \centering
    \vspace{-0.2cm}
    \includegraphics[width=0.7\linewidth]{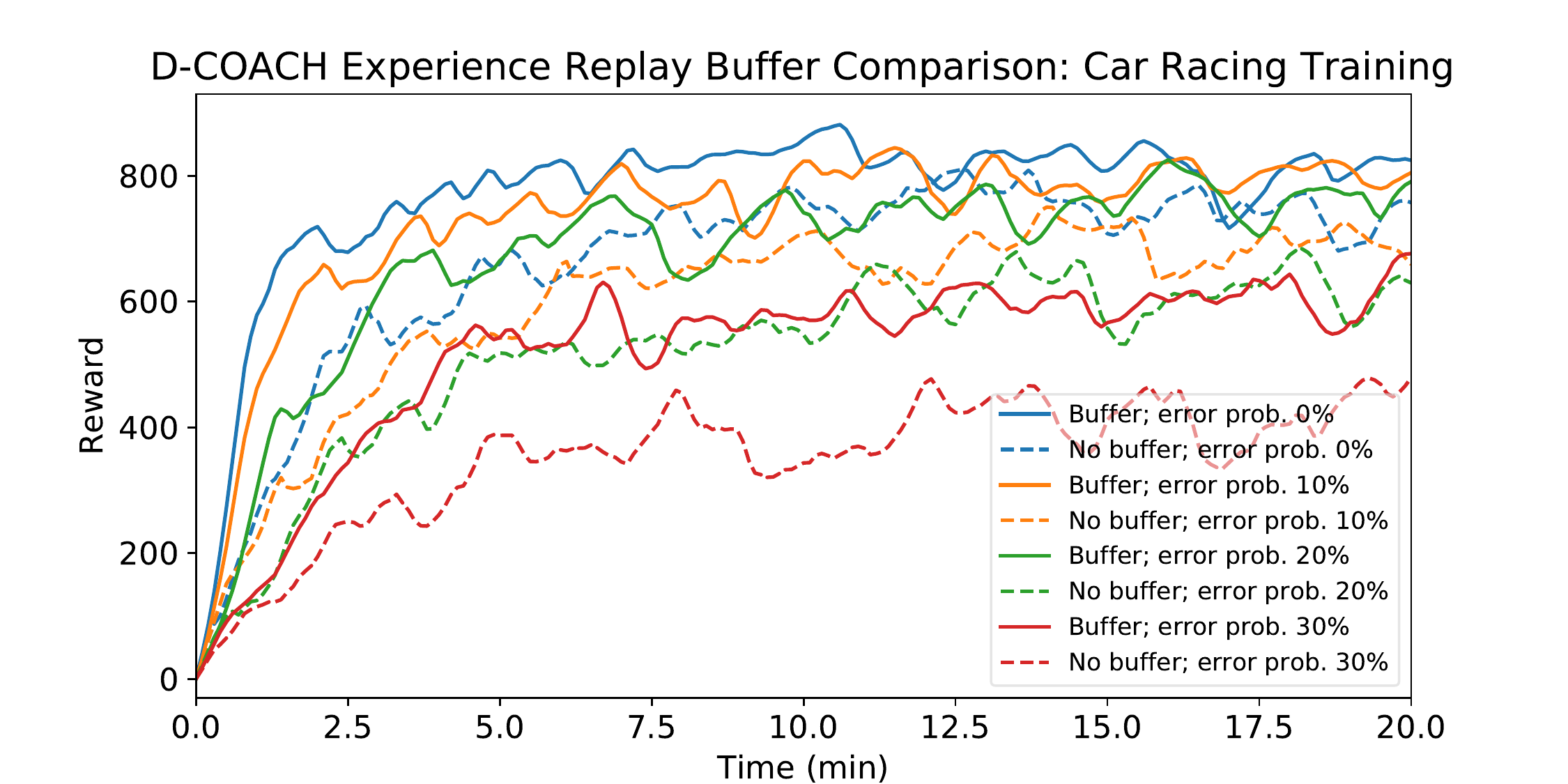}
    \vspace{-0.2cm}
    \caption{Comparison between using or not experience replay buffer for different values of $P_\mathit{err}$ in the CarRacing problem. Buffer: $K = 1000$; $b = 10$; $N = 100$. $P_{h}$: $\alpha = 0.6$; $\tau = 0.000015$. Simulated teacher network learning rate: $0.0003$.}
    \label{fig:buffer_car_racing}
\end{figure}

A comparison of D-COACH with and without the use of an experience replay buffer is carried out by means of the simulated teacher. To test the behavior of these scenarios when erroneous feedback is added, different values of $P_{\mathit{err}}$ are selected. These results can be seen in \figurename~{\ref{fig:buffer_cart_pole}} and \figurename~{\ref{fig:buffer_car_racing}} (for better readability, no confidence intervals were added).

In \figurename~{\ref{fig:buffer_cart_pole}} and \figurename~{\ref{fig:buffer_car_racing}} the learning curves show a large difference between the processes of learning that use experience replay buffer with respect to the cases without the buffer. In the case without the buffer, which is more similar to the original COACH, it is possible to see that the learning agent is not benefiting from the advised corrections as much as it can do when the pieces of advice are kept in the memory. For instance, we can see that D-COACH learns more from corrections with $20 \%$ of mistakes when using the buffer than in the case of perfect corrections, but without any buffering. This means the buffer is necessary for increasing the use of the information available, even when this information is corrupted and not clean.
\vspace{-0.4cm}

\subsection{Comparison of DRL and D-COACH using real human teachers}
\begin{figure}[t]
    \centering
    \vspace{-0.2cm}
    \includegraphics[width=0.7\linewidth]{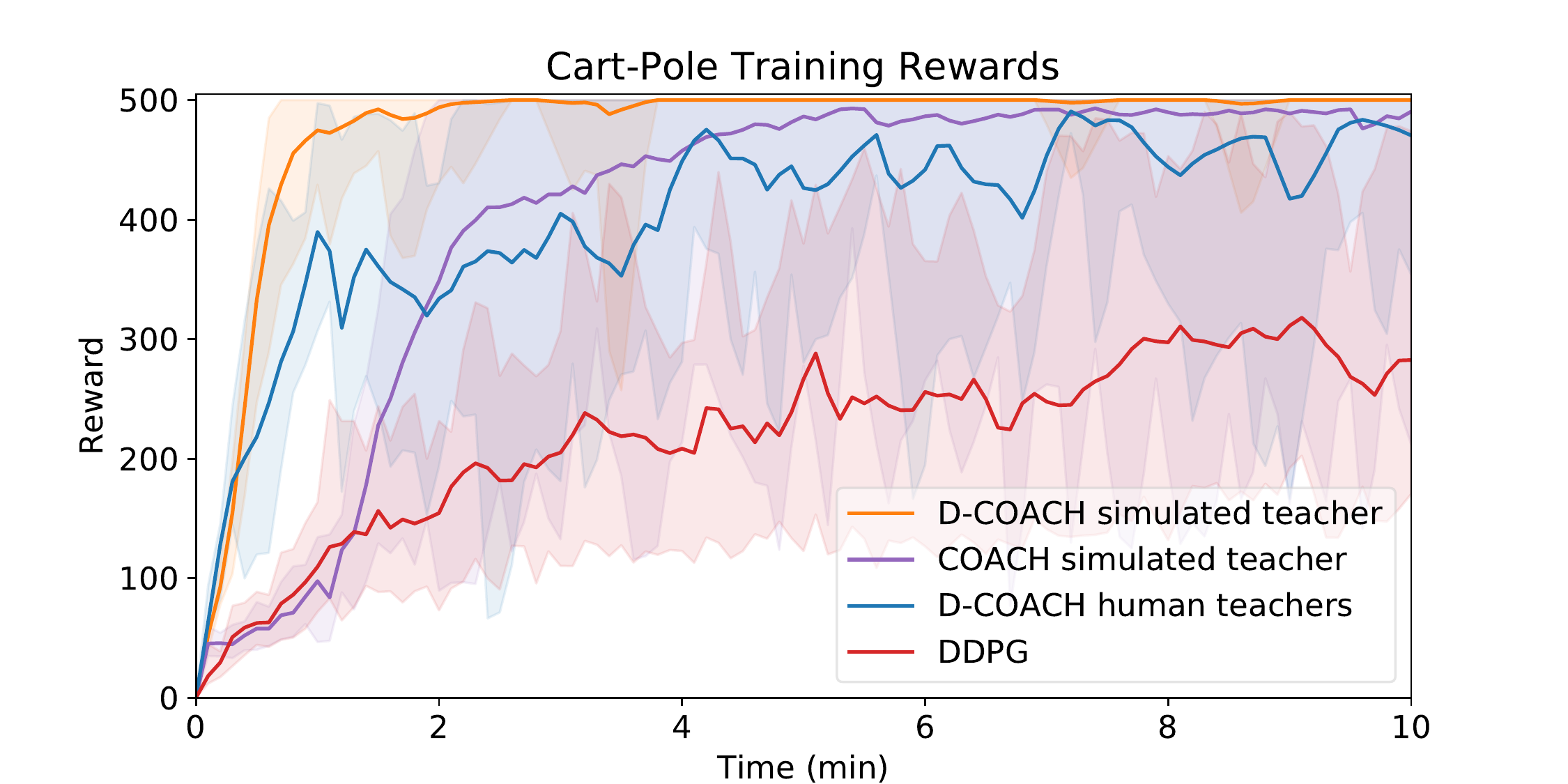}
    \vspace{-0.2cm}
    \caption{Cart-Pole training. Buffer: $K = 200$; $b = 10$; $N = 50$. $P_{h}$: $\alpha = 0.6$; $\tau = 0.0003$. Human teacher network learning rate: $0.003$; Simulated teacher network learning rate: $0.0003$.}
    \label{fig:cartpole_results}
\end{figure}

These experiments are intended to compare the learning process of D-COACH (simulated teacher and human teacher) with the DRL algorithm DDPG. Taking into account that the Cart-Pole problem has a low dimensional state space, the original COACH, based on basis functions, is also included in the comparison. In this case, $P_\mathit{err}=0\%$ was used for the simulated teachers. The results of this problem are shown in \figurename~{\ref{fig:cartpole_results}}, wherein it is possible to see that COACH-like methods outperform the DRL agent with a large difference. When using the simulated teacher, D-COACH learns faster than the original COACH. The performance of D-COACH with human teachers decreases with respect to the simulated teacher. This is because human teachers are not perfect and make mistakes, but they are being compared with a simulated teacher with $P_\mathit{err}=0\%$, which means that it makes no mistakes. Also because the simulated teacher model is quite simple to represent the complexity of the human behavior, then, although it is not very realistic, it is still useful for comparisons of interactive learning strategies under similar conditions.

\begin{figure}[t]
    \centering
    \vspace{-0.2cm}
    \includegraphics[width=0.7\linewidth]{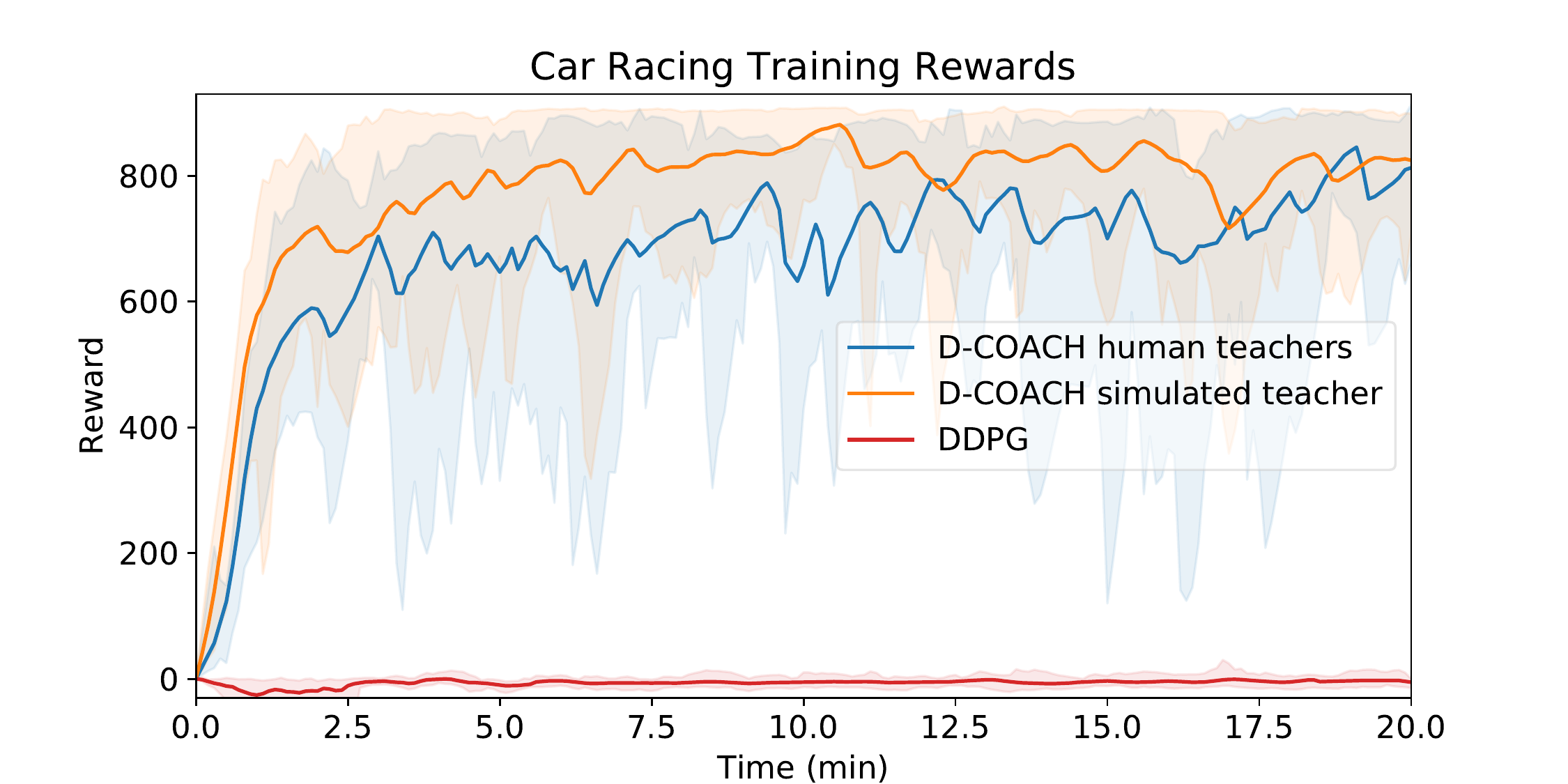}
    \vspace{-0.2cm}
    \caption{Racing Car training. Buffer: $K = 1000$; $b = 10$; $N = 100$. $P_{h}$: $\alpha = 0.6$; $\tau = 0.000015$. Human teacher network learning rate: $0.001$; Simulated teacher network learning rate: $0.0003$.}
    \label{fig:racing_car_results}
\end{figure}

In \figurename~{\ref{fig:racing_car_results}} the learning curves of the Car Racing problem are presented. Again, D-COACH results in a fast convergence. Unlike reported results of DRL algorithms for this problem, in the very early minutes D-COACH reaches high performance policies that have not been obtained by most of the DRL approaches, to the best of our knowledge. If we compare a policy trained with D-COACH for approximately 75 minutes by an experienced teacher against several state-of-the-art DRL approaches, it can be seen that it outperforms most of them (see Table \ref{CarRacing_table}). The problem is considered to be solved if the agent gets an average score of 900 or more over 100 random tracks. However, we observed that this value can substantially vary between different evaluations, so in Table \ref{CarRacing_table}, the obtained range of values over 20 evaluations is presented for D-COACH.
\vspace{-0.4cm}

\begin{table}[t]
\centering
\caption{Car Racing state-of-the-art learning algorithms comparison. DRL results taken from \cite{Ha2018}.}
\label{CarRacing_table}
\begin{tabular}{lc}
\multicolumn{1}{c}{\textbf{Method}}      & \multicolumn{1}{l}{\textbf{Average Score over 100 Random Tracks}} \\ \hline\hline
DQN                                      & 343 $\pm$ 18                                                      \\ \hline
A3C (continuous)                         & 591 $\pm$ 45                                                      \\ \hline
A3C (discrete)                           & 652 $\pm$ 10                                                      \\ \hline
ceobillionaire’s algorithm (unpublished) & 838 $\pm$ 11                                                      \\ \hline
Full World Model                         & 906 $\pm$ 21                                                      \\ \hline
\textbf{D-COACH (experienced teacher)}                         & \textbf{895 - 909 $\pm$ 18 - 80} \\
& \textbf{Average over 20 evaluations: 903 $\pm$ 46}
\\ \hline
\end{tabular}
\end{table}

\subsection{Validation in a real system}
In the third problem that we called Duckie Racing, an agent has to learn to drive a Duckiebot (from the project  Duckietown \cite{Paull2017} with modifications from the Chile Duckietown Team\footnote{\url{https://github.com/Duckietown-Chile/}}) autonomously through a track based on raw visual information of an onboard camera. The actions in this problem are the forward velocity and the steering angle of the Duckiebot. Two tasks are set for this environment: (i) driving the Duckiebot freely through the track, with permission to drive in both lanes, and (ii) driving the Duckiebot only in the right lane, which demands more accuracy in driving. In this problem, an episode stops if the robot leaves the track/right lane, or after 30 seconds. The performance index in this task is the percentage of the total track length traveled during the episode. Hence the faster and more accurate the Duckiebot drives, the more distance it will travel.

\begin{figure}[t]
    \centering
    \begin{minipage}{.5\textwidth}
    \vspace{-0.2cm}
    \includegraphics[width=1.0\linewidth]{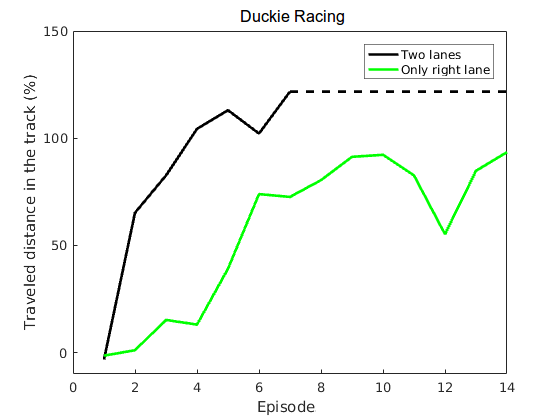}
    \vspace{-0.2cm}
    \caption{Duckie Racing training.}
    \label{fig:racing_duckie_results}
    \end{minipage}%
    \begin{minipage}{.5\textwidth}
    \centering
    \includegraphics[width=1.0\linewidth]{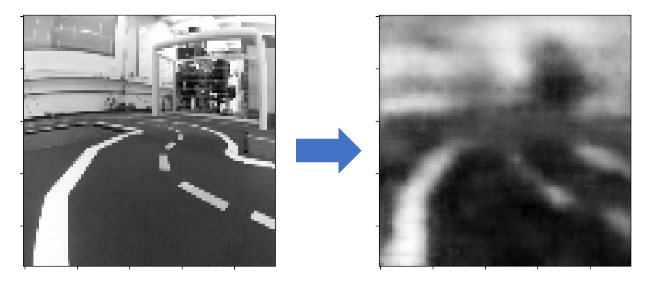}
    \vspace{-0.2cm}
    \caption{Duckie Racing autoencoder input (left) vs output (right).}
    \label{fig:AE_duckie}
    \end{minipage}
\end{figure}

This problem is not used for comparisons of the methods, but only as a validation of D-COACH using experience replay, which showed to be the best alternative in the previous problems. \figurename~\ref{fig:racing_duckie_results} shows the learning curve for each of the tasks explored in this environment with a real robot and a real human teacher. The curves and the video\footnote{https://youtu.be/vcEtuRrRIe4} attached to this paper show that the system quickly learns to drive properly through the road based only on the human corrections. As expected, the policy is faster when the robot has the freedom to drive over both lanes. Learning this task with RL would definitely take more training time, and might need an external perception system to compute the reward function, whereas with D-COACH this performance index does not have any influence on the learning process, rather it is used for descriptive and comparative purposes.

\section{Conclusions}

This work presented D-COACH, an algorithm for training policies modeled with DNNs interactively with corrective advice. The method was validated in a problem of low-dimensionality, along with problems of high-dimensional state spaces like raw pixel observations, with a simulated and a real robot environment, and also using both simulated and real human teachers. 

The use of the experience replay buffer (which has been well tested for DRL) was re-validated for this different kind of learning approach, since this is a feature not included in the original COACH. The comparisons showed that the use of memory resulted in an important boost in the learning speed of the agents, which were able to converge with less feedback, and to perform better even in cases with a significant amount of erroneous signals.  

The results of the experiments show that teachers advising corrections can train policies in fewer time steps than a DRL method like DDPG. So it was possible to train real robot tasks based on human corrections during the task execution, in an environment with a raw pixel level state space. 
The comparison of D-COACH with respect to DDPG, shows how this interactive method makes it more feasible to learn policies represented with DNNs, within the constraints of physical systems. DDPG needs to accumulate millions of time steps of experience in order to obtain good performances as shown in \cite{Lillicrap2015}. However, this is not always possible with real systems.



\begin{acknowledgement}
This work was partially funded by FONDECYT Project 1161500. A portion of it has taken place in the  University of Chile Duckietown's headquarters, \emph{FabLab U. de Chile} (\url{http://www.fablab.uchile.cl/}). Special thanks to Matias Mattamala, who provided the necessary tools to do the tests with the Duckiebots.
\end{acknowledgement}

\bibliographystyle{IEEEtran}
\bibliography{author}
\end{document}